\documentclass[sigconf]{acmart}
\AtBeginDocument{%
  }

\usepackage{algorithm}
\usepackage{algpseudocode}
\usepackage{multirow}
\usepackage{makecell}
\usepackage{subfig}
\usepackage{float}

\def\methodname{MatTA }

\copyrightyear{2025}
\acmYear{2025}
\setcopyright{cc}
\setcctype{by}
\acmConference[KDD '25]{Proceedings of the 31st ACM SIGKDD Conference on Knowledge Discovery and Data Mining V.2}{August 3--7, 2025}{Toronto, ON, Canada}
\acmBooktitle{Proceedings of the 31st ACM SIGKDD Conference on Knowledge Discovery and Data Mining V.2 (KDD '25), August 3--7, 2025, Toronto, ON, Canada}
\acmDOI{10.1145/3711896.3737245}
\acmISBN{979-8-4007-1454-2/2025/08}

\begin{document}

\title{Matryoshka Model Learning for Improved Elastic Student Models} 

%


\author{Chetan Verma}
\authornote{Indicates equal contribution}
\affiliation{\institution{Google}\city{Mountain View}\state{CA}\country{USA}}
\email{ckverma@google.com}

\author{Aditya Srinivas Timmaraju}
\affiliation{\institution{Google DeepMind}\city{Bengaluru}\state{KA}\country{India}}
\email{adityasrinivas@google.com}
\authornotemark[1]

\author{Cho-Jui Hsieh}
\affiliation{\institution{Google}\city{Los Angeles}\state{CA}\country{USA}}
\email{cjhsieh@google.com}

\author{Suyash Damle}
\affiliation{\institution{Google}\city{Mountain View}\state{CA}\country{USA}}
\email{suyashd@google.com}

\author{Ngot Bui}
\affiliation{\institution{Google}\city{Mountain View}\state{CA}\country{USA}}
\email{bpngot@google.com}

\author{Yang Zhang}
\affiliation{\institution{Google}\city{Mountain View}\state{CA}\country{USA}}
\email{zhangya@google.com}

\author{Wen Chen}
\affiliation{\institution{Google}\city{Mountain View}\state{CA}\country{USA}}
\email{chenwen@google.com}

\author{Xin Liu}
\affiliation{\institution{Google}\city{Mountain View}\state{CA}\country{USA}}
\email{xinliujune@google.com}

\author{Prateek Jain}
\affiliation{\institution{Google DeepMind}\city{Bengaluru}\state{KA}\country{India}}
\email{prajain@google.com}

\author{Inderjit Dhillon}
\affiliation{\institution{Google}\city{Mountain View}\state{CA}\country{USA}}
\email{isd@google.com}

\renewcommand{\shortauthors}{Chetan Verma et al.}


\ccsdesc[500]{Computing methodologies~Neural networks}
\ccsdesc[500]{Computing methodologies~Ranking}


\keywords{Online distillation, Elastic inference, Matryoshka representations}


\begin{abstract}

Industry-grade ML models are carefully designed to meet rapidly evolving serving constraints, which requires significant resources for model development. In this paper, we propose MatTA, a framework for training multiple accurate Student models using a novel Teacher-TA-Student recipe. TA models are larger versions of the Student models with higher capacity, and thus allow Student models to better relate to the Teacher model and also bring in more domain-specific expertise. Furthermore, multiple accurate Student models can be extracted from the TA model. Therefore, despite only one training run, our methodology provides multiple servable options to trade off accuracy for lower serving cost. We demonstrate the proposed method, MatTA, on proprietary datasets and models. Its practical efficacy is underscored by live A/B tests within a production ML system, demonstrating 20\% improvement on a key metric. We also demonstrate our method on GPT-2 Medium, a public model, and achieve relative improvements of over 24\% on SAT Math and over 10\% on the LAMBADA benchmark.

\end{abstract}
\maketitle

\section{Introduction}
\begin{figure*}[h!]
    \centering
    \subfloat[\centering  ]{{\includegraphics[width=0.4\textwidth]{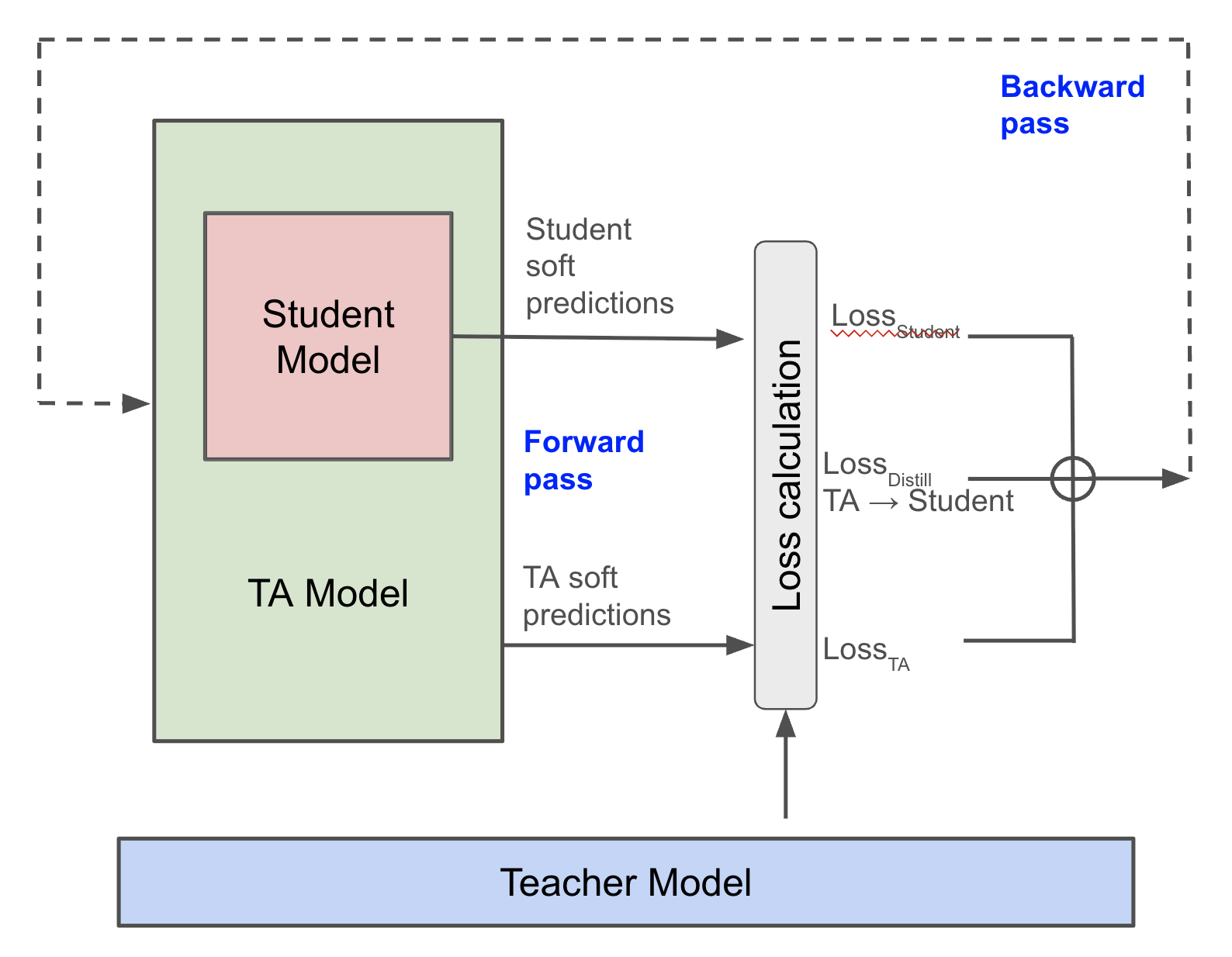} }}%
    \qquad
    \subfloat[\centering]{{\includegraphics[width=0.4\textwidth]{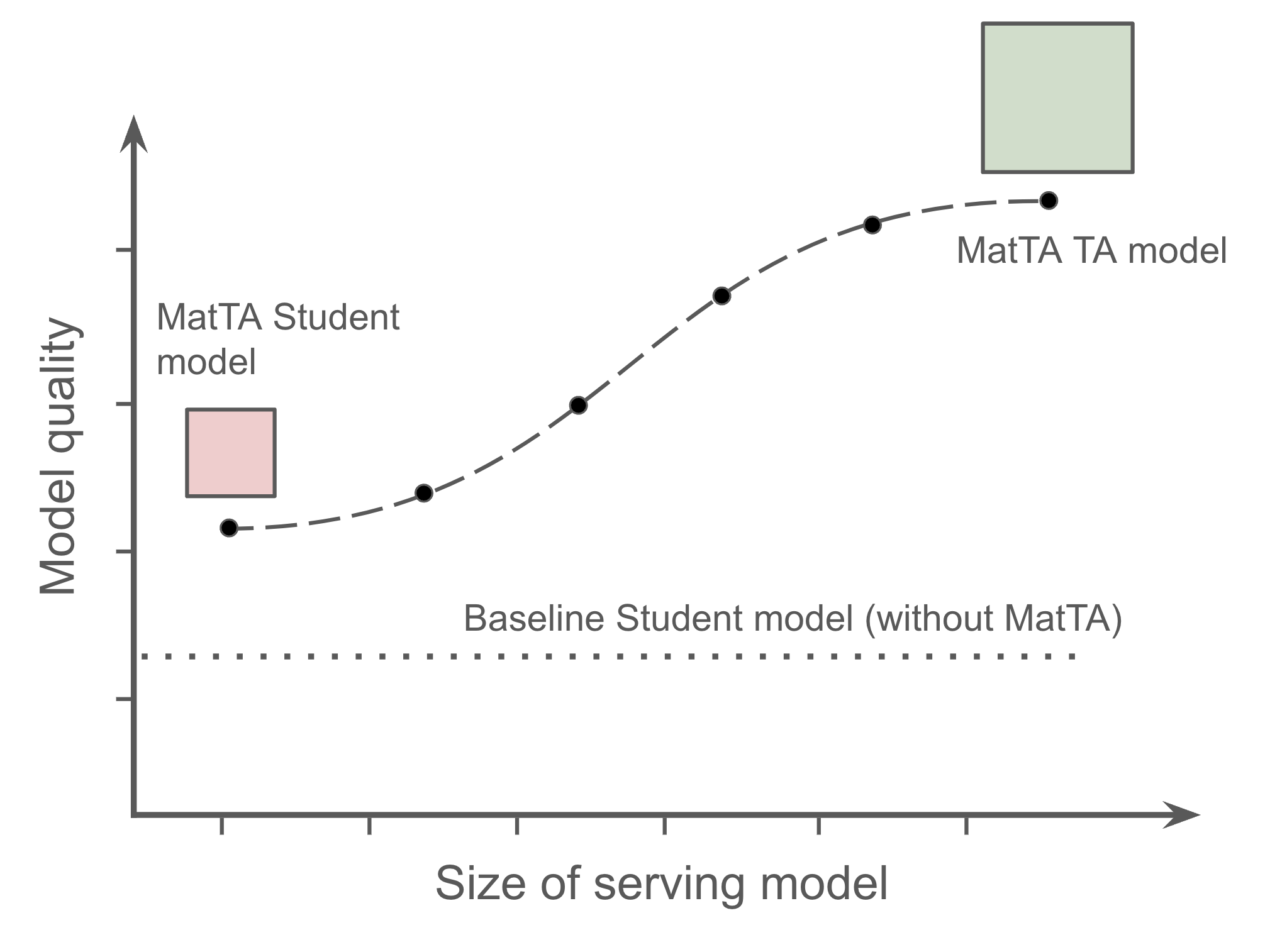} }}%
    \caption{(a) MatTA: a novel elastic distillation framework. In order to generate elastic servable Student models from a single training run, we create a Teaching Assistant~(TA) model from the given serving model. TA and Student are co-trained with Student learning from original (Teacher's) labels as well as distilling from TA. (b) MatTA can extract a range of compelling Student models, each of which surpass an independently trained original Student model. Experiments show significant improvement even while retaining identical architecture.}%
    \label{fig:example}%
\end{figure*}

Industrial applications employ a spectrum of Machine Learning~(ML) models to serve predictive tasks, ranging from ranking and retrieval to classification and content tagging. Serving models are typically optimized for their quality, given their causal impact on key business metrics. At the same time, these models must satisfy practical requirements imposed by the overall infrastructure. These requirements include size constraints due to serving hardware limitations (memory, bandwidth, and latency), as well as the need to make real-time, high-quality predictions based on limited amounts of training data. Unsurprisingly, in contrast to foundational Large Language Models (LLMs) that span hundreds of billions of parameters, a significant number of serving production models in the real world are much smaller in size.

Typically, such serving-sized models are: (a) trained from scratch using application-specific data; (b) imbued with "world knowledge" or more refined labels using distillation from either a larger model of similar architecture or from a much larger foundational Teacher model \cite{li2023prompt,yuan2020parameter}. However, such approaches routinely run into three practical challenges.

First, the training procedures for such servable models are rigid: they produce a single model from a training run. At the same time, model requirements are elastic, owing to the heterogeneity in serving infrastructure and the constant evolution of hardware and user behavior. ML teams need to periodically retrain their models to use the latest hardware efficiently. This is resource intensive and demands significant model development costs. Moreover, the rapid pace of development in accelerator technology has further fueled this problem.
Second, the pre-trained foundational Teacher models that are used in the industry are extremely large (eg. 405B parameters for Llama 3.1 \cite{llama2024}) and, hence, not "relatable" when compared to the serving Student models (usually in the range of 10s-100s of million parameters). Furthermore, the Teacher models might not even consume the same set of features as the application-specific serving model and would require extensive fine-tuning before they can be used for distillation. Third, since the amount of application-specific data can be limited, careful selection of the optimizer and pairing it with the correct model architecture is critical.

In this paper, we present Matryoshka Teaching Assistant~(MatTA), a novel elastic distillation framework designed to address the aforementioned challenges. MatTA (as shown in Figure~\ref{fig:example}(a)) leverages prior work in Matryoshka models~\cite{mrl-2022,devvrit2024matformer,cai2024flextron,wang-etal-2024-train} to create a novel Teacher-TA-Student distillation method. In this method, the Student models are nested within the TA~(Teaching Assistant) model. Specifically, we train a large TA model within which a servable Student model is nested—similar to the Matformer or Flextron architectures~\cite{devvrit2024matformer, cai2024flextron}. Logits from the Teacher model—which can be a large foundational model or an independently trained large model—are distilled into the TA and Student models. The Student improves by learning from a more relatable TA. We train this ensemble using Shampoo \cite{shampoo}, a second-order optimizer well-suited for both the Matryoshka structure of TA-Students and for situations with limited data.

The MatTA framework offers three key advantages:

a. As observed in existing papers on nested models~\cite{devvrit2024matformer,cai2024flextron}, we can mix-and-match different layers/parts of the model to produce multiple Student models; see Figure~\ref{fig:example}(b). These multiple servable Students provide different cost-quality trade-offs, enabling the selection of an appropriate student model based on the latest hardware/serving configurations and product requirements.

b. Due to distillation from the TA model into the Student model(s), the learned Students tend to be more accurate than traditional stand-alone models learned via Teacher-Student distillation (see Section~\ref{sec:exps}).

c. The second-order Shampoo optimizer, along with regularization afforded by the Matryoshka structure, allows MatTA to learn accurate models despite limited application-specific data.

While Matryoshka structures and multi-level distillation have been explored to some degree in the literature, the key novelty of our work lies in formulating and demonstrating the hypothesis that a Matryoshka structure, combined with a novel Teacher-TA-Student distillation framework, can provide significantly more accurate and "elastic" servable models—a critical requirement for industry-grade ML applications. See Section~\ref{sec:related} for further details.

We demonstrate the effectiveness of MatTA on a publicly available, serving-sized model: GPT-2 Medium~\cite{gpt2}. We show that our method can improve the original GPT-2 model's performance on popular benchmarks such as SAT Math and LAMBADA by a relative margin of 24\% and 10\%, respectively. In addition, we show that a range of models can be extracted using Mix`n'Match~\cite{devvrit2024matformer} on the MatTA GPT-2 models, offering competitive model quality. We also report results on multiple proprietary industry applications and demonstrate that MatTA indeed makes a significant difference for industry-grade servable models. In particular, we demonstrate that MatTA can improve AUROC for a critical industry application by 8\%, and the improved offline loss metrics translate to a gain of 20\% in a key metric during live experiments.

We also conduct multiple ablation studies to demonstrate how different components of MatTA interact and contribute to improvements in model quality. In particular, we provide empirical evidence of how Teacher-TA-Student distillation improves model quality. Furthermore, we demonstrate that second order optimizers, such as Shampoo, can exploit the proposed nested architecture in MatTA, providing super-additive quality improvements.

The remainder of this paper details past related work, the \methodname methodology, experimental setup, results, and ablation studies, providing a comprehensive evaluation of our proposed approach and its practical implications.

\section{Related Work}\label{sec:related}

{\bf Knowledge Distillation.}
Knowledge distillation (KD) is a commonly used technique to improve the quality of a smaller student model using fine-grained knowledge stored in larger teacher models ~\cite{distillation,dehghani2023scaling,li2022blip,touvron2021training,fedus2022switch}. Researchers have also studied generalization methods to improve a student model by using knowledge within the network itself~\cite{zhang2019your,lan2019self}. Today, KD is widely used across different domains, including speech recognition \cite{markov2016robust,kurata2020knowledge,liu2019end}, computer vision~\cite{yang2023does,Puy_2024_CVPR,li2021align} and Large Language models~(LLMs)~\cite{pmlr-v202-liang23j,wu2021one}. 

In production ML systems, KD is commonly achieved by using the class probabilities of the teacher model as ``soft targets” ~\cite{distillation}, or by using intermediate representations learned by the teacher~\cite{romero2014fitnets}. While intermediate representation distillation based approaches seem promising, so far their impact has been relatively less, and in this paper, we mainly focus on soft logits distillation. 
While a majority of past KD research focuses on a two staged approach where a pre-trained teacher model is used to improve student models, work in \cite{zhang2018deep,onlinedistillation} explores training these models together via online distillation. In industrial settings where the underlying patterns are non-stationary due to recency and seasonality biases, a significant disadvantage of traditional two-staged distillation is that it adds additional delay to the model that is used to serve production requests. This is because the teacher model first needs to be trained on the latest data, after which the student model is trained. In addition, online distillation does not require additional infrastructure for storage of Teacher's logits, as is typically done with traditional two staged distillation. MatTA uses online distillation to improve serving model quality. 
Some recent research has explored multi-stage distillation, for example, Quill~\cite{quill} has a Professor-Teacher-Student setup where the Professor is a powerful LLM with RAG while Teacher-Student form the typical distillation setup. In contrast, MatTA introduces nesting of the TA and Student models which not only learns a more accurate Student model but also produces elastic Student models.

{\bf Elastic Inference.}
The increasing diversity of deployment platforms, from high-end servers to resource-constrained edge devices, coupled with the substantial cost of model training has motivated recent research in elastic inference \cite{cai2019once,zhang2019your,devvrit2024matformer,cai2024flextron,valipour2023sortednet}. \cite{cai2019once,zhang2019your} initiated research in the domain of generating a range of models of varying sizes from a single training run with CNNs being the base model. Recently,~\cite{devvrit2024matformer} generalized the approach to LLMs and demonstrated that multiple LLMs can be extracted using Mix`n'Match. The paper~\cite{cai2024flextron} proposes a post-training method to extract multiple models from a trained LLM and introduces routing mechanisms to choose between them dynamically. 
In contrast to the above research, our paper focuses on starting from a given production model and introduces a larger ``TA'' model by \textit{M-nesting} (defined in Section~\ref{sec:architecture}) selected layers or the entirety of the original model. We also show how co-training the original Student model with the TA leads to a significantly more accurate production Student model along with a range of high quality models via Mix`n'Match. In addition, our focus is on models that are much smaller than foundational models that \cite{devvrit2024matformer, cai2024flextron} focus on. To the best of our knowledge, this is the first work demonstrating elastic techniques at 10-100M model scale.

{\bf Second Order Optimizers.}
First-order optimizers, such as Adagrad~\cite{duchi2011adaptive} and Adam~\cite{kingma2014adam}, are widely used for training neural networks. While effective, these methods employ adaptive learning rates for each coordinate independently, neglecting potential correlations between them.  Full-matrix Adagrad, which maintains the full second-moment matrix ($\sum_t g_t g_t^T$), can capture these higher-order correlations but suffers from prohibitive computational cost due to the quadratic size of the matrix.  To address this, methods like Shampoo~\cite{gupta2018shampoo} and K-FAC~\cite{martens2015optimizing} approximate the second-moment matrix using the Kronecker product of two smaller matrices.  Shampoo has demonstrated effectiveness in various real-world applications~\cite{anil2022factory,anil2020scalable}, and several improvements have been proposed~\cite{duvvuricombining,vyas2024soap}. In this work, we empirically demonstrate the effectiveness of higher-order optimizers for training Student and TA models using the proposed Matryoshka structure.

\section{Methodology}
Without loss of generality, we describe our method \methodname using a setup where a Teacher model generates training labels for a small Student model which is used for serving. While this is a commonly used approach in industry, the discussion below applies even if the labels are from events (such as clicks, purchases) and not from a Teacher model.

The original (Student) model is \textit{M-nested} to create a larger Teaching Assistant (TA) model. The Student and TA models are trained concurrently via online distillation ~\cite{onlinedistillation}. As shown in Algorithm ~\ref{alg:tagl_algorithm}, for each mini-batch of the training data, a composite loss function $L$ is created. The loss function is detailed in Section ~\ref{sec:Loss}. We use parameter sharing between the Student and the TA models such that the latter comprises a strict superset of the parameters of the former. Section ~\ref{sec:architecture} outlines different approaches to \textit{M-nestify} a given servable Student model to create the TA model. This is done by in turn \textit{M-nesting} a chosen set of layers in the original model. Alternatively, a given model can also be \textit{M-nested} by increasing the depth of the model i.e., adding newer layers that are unique to the TA model, or by increasing the dimensionality  $d\_model$ of the model. In Section~\ref{sec:architecture}, we provide examples for \textit{M-nesting} commonly used layers in modern neural networks. Note that varying the number of hidden nodes in the feed forward module of a Transformer block ~\cite{transformer} as shown in ~\cite{devvrit2024matformer} is one approach to do this.
We refer the reader to \cite{onlinedistillation} for more details on online distillation.

\algrenewcommand\algorithmicrequire{\textbf{Inputs:}}
\algrenewcommand\algorithmicensure{\textbf{Outputs:}}

\begin{algorithm}
\caption{Matryoshka Teaching Assistant (MatTA)}\label{alg:tagl_algorithm}
\begin{algorithmic}
\Require \\
$\Theta$: Student model parameter set \\
$\Phi$: TA model parameter set: $\Theta \subset \Phi$ \\
$\omega \in \mathcal{R}^3$: Loss weights
\Ensure \\
$\{M\}$: set of models with varying sizes where $M:=\{ \theta : \theta \subset  \Phi \}$;
\While{$not\ converged$}
\State $y$: Labels (from a Teacher model or from events)
\State $p_S$: Student model predictions
\State $p_{TA}$: TA model predictions \\
\State $L_S = f(p_S, y)$ // Student loss
\State $L_{TA} = f(p_{TA}, y)$ // TA loss
\State $L_D = g(p_S, p_{TA})$ // Distillation loss
\State $L = \omega^T * [L_S, L_{TA}, L_D]$ // Form composite loss
\State $\Theta \leftarrow \Theta - \eta\nabla_\Theta{L}$ // Update student parameters
\State $\Phi \leftarrow \Phi - \eta\nabla_\Phi{L}$ // Update teacher parameters
\EndWhile \\

$\{M\} \leftarrow $ Mix`n'Match($\Theta, \Phi$)
\end{algorithmic}
\end{algorithm}

\subsection{Loss}
\label{sec:Loss}
As shown in Algorithm~\ref{alg:tagl_algorithm}, the parameters $\Theta$ and $\Phi$, corresponding to the Student model and the TA model respectively, are updated together during training. Let us consider a multi-class classification problem. The per-example losses corresponding to these models are $L_S$ and $L_{TA}$, and are defined as

\begin{eqnarray}
L_S & = & - \Sigma_{i \in C}  \,\,\, y_i * \log(p_{S, i}), \;\; {\mbox{\rm and}} \\
L_{TA} & = & - \Sigma_{i \in C}  \,\,\, y_i * \log(p_{TA, i}).
\end{eqnarray}

Here $C$ refers to the set of classes and $y_i$ are the ground truth probabilities for class $i$ from the Teacher model. The quantities $p_{S, i}$ and $p_{TA, i}$ refer to the predicted probabilities by the Student and TA models respectively for class $i$. $L_S$ and $L_{TA}$ capture the losses of the two models' predictions as compared to the ground truth. We introduce a third distillation based loss, $L_D$ to distill knowledge from the TA to the Student model, which is defined per example as: 
\begin{equation}
    L_D = - \Sigma_{i \in C}   \,\,\,  p_{TA, i} * \log(p_{S, i}).
\end{equation}

This is based on the observation that the higher capacity TA model reaches better model quality compared to the original model. At the same time, the TA model has a high "relatability" with the Student model as they have comparable sizes, architectures and features. So, the Student model can benefit via distillation from the predictions of the TA model. During the backward pass, we disallow the gradient to update the TA model parameters $\Phi$ based on $L_D$. The loss that we optimize for is a composite loss $L$ which is a weighted combination of $\{L_S, L_{TA}, L_D\}$ using weights $\omega$, which are tuned as a hyperparameter.

\subsubsection{Training curriculum}

The TA model and the Student model both have poor quality in the beginning of the training. As a result, it helps to discard the contribution of $L_D$ in $L$ for the initial steps of the training. We achieve this by introducing a hyperparameter to control the ramp up duration of the contribution of $L_D$ in $L$. 

\subsection{Architecture}
\label{sec:architecture}
An integral component of the \methodname approach is parameter sharing between the Student and the TA models. The TA model is intended to have higher capacity i.e., more trainable parameters than the Student model. We define \textit{Matryoshka nesting} or \textit{M-nesting} in short, to be the expansion of an original Student model to create a corresponding TA model. M-nesting can be achieved either layer wise, depth wise or by a combination of both. 

\subsubsection{Layer M-nesting}
\label{sec:layer_expansion}
At a high level, layer wise \textit{M-nesting} increases the size of one or more dimensions of the multi-dimensional weight parameters of a layer. \cite{devvrit2024matformer,cai2024flextron} describe one way to do this by increasing the layer width. It is important to note that after implementing \textit{M-nesting} on a layer, there are two sets 
of activations at the output of the layer instead of one. One set of activations corresponds to the Student model, and the other to the TA model. The example below shows the explicit calculations required to \textit{M-nestify} a simple Dense layer in this general case of two inputs.

\textbf{Example: M-nesting a Dense Layer}

\begin{figure*}
    \centering
    \includegraphics[width=0.2\textwidth, angle=90]{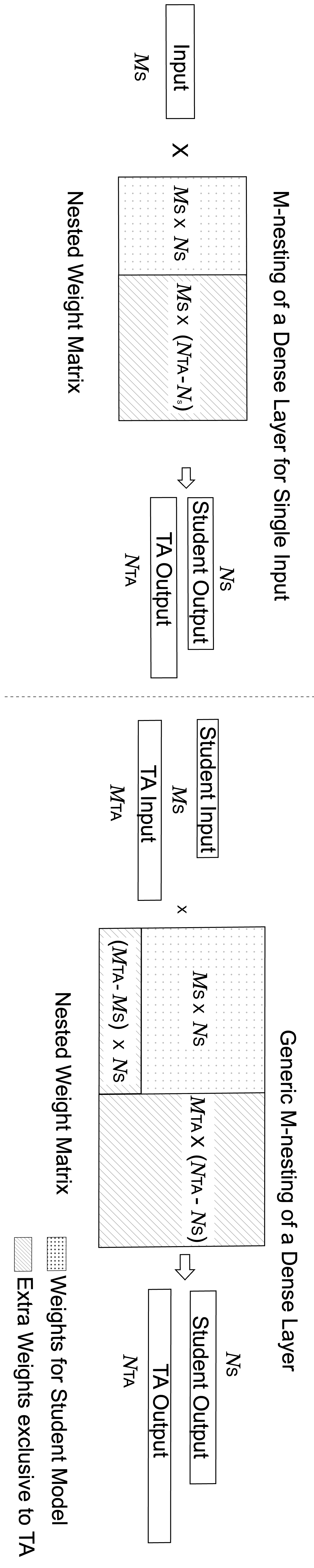}
    \caption{M-Nesting a Dense layer. {\bf (Left)}  M-Nested Dense layer with one input. This is the situation when this layer is the first one to get M-nested. {\bf (Right)} M-Nested Dense layer with two inputs. This is the general case throughout the rest of the model after at least one layer has already been M-nested.}
    \label{fig:kdd_dense}
\end{figure*}

 Algorithm~\ref{alg:matformed_dense_algorithm} shows how a Dense layer would be \textit{M-nested} in the general case that there are two inputs; one corresponding to the Student, and one to the TA model. The layer parameters ${W}_S$ for the Student model are a subset of the parameters ${W}_{TA}$ which correspond to the TA. Figure~\ref{fig:kdd_dense} shows this visually.

\textbf{Example: M-nesting an arbitrary layer}

\begin{figure}[htb!]
    \centering
    \includegraphics[width=0.5\textwidth]{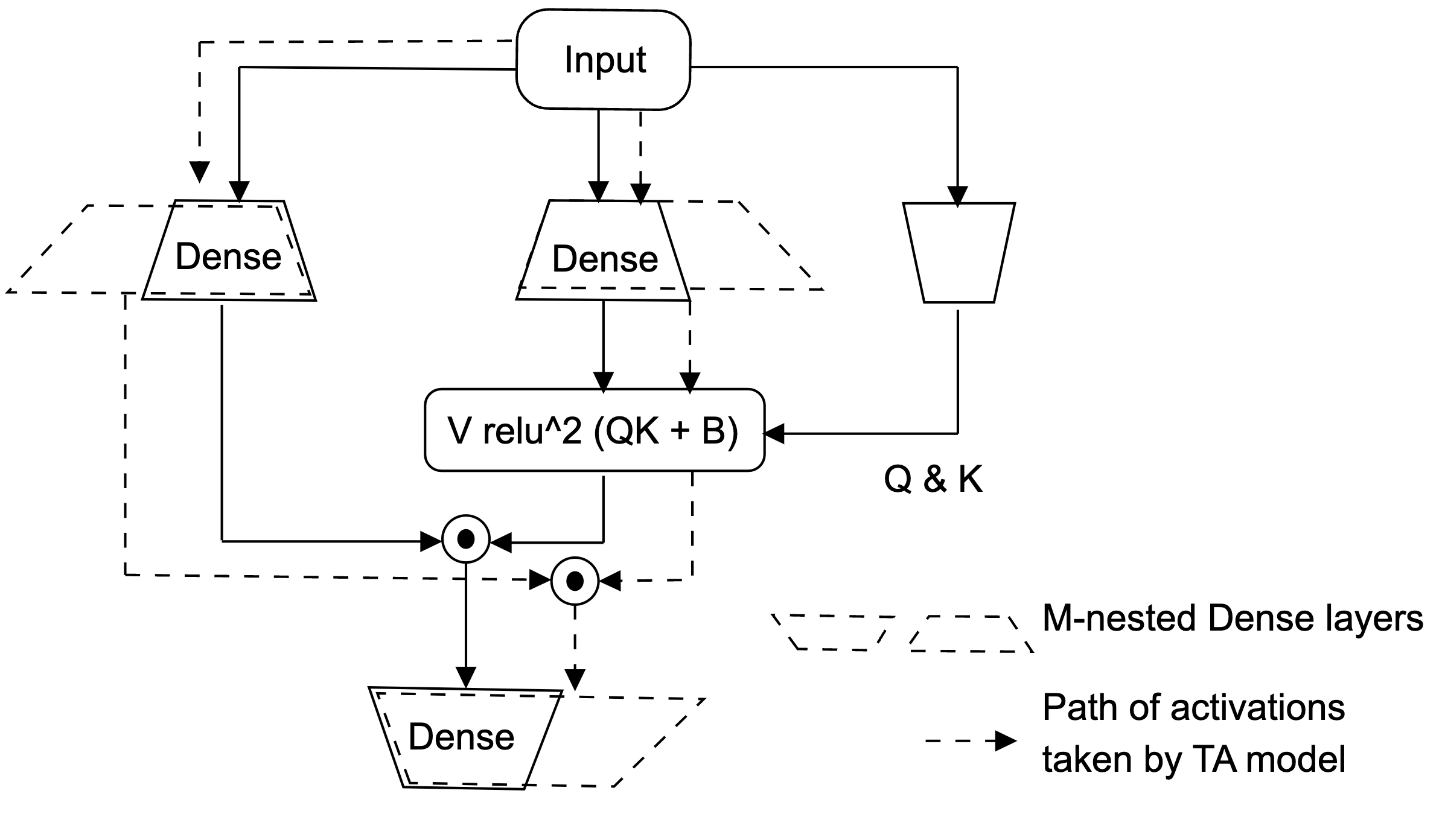}
    \caption{M-nesting of a GAU unit~\cite{gauunit}}
    \label{fig:gau_tagl}
\end{figure}

The procedure for \textit{M-nesting} an arbitrary layer is straightforward: find projection matrices used within the layer, and have them undergo \textit{M-nesting} similar to a Dense layer above. Figure~\ref{fig:gau_tagl} shows an example of how a Gated Attention Unit (GAU) \cite{gauunit} can be M-nested to create a corresponding GAU layer for the TA model. All occurences of a Dense layer in Figure~\ref{fig:gau_tagl} are replaced with an M-nested Dense layer to M-nestify the GAU unit.

 \algnewcommand\algorithmic {\textbf{Outputs:}}

\begin{algorithm}
\caption{M-nesting a Dense layer}\label{alg:matformed_dense_algorithm}
\begin{algorithmic}
\Require \\
$M_{S}$: Size of input to Student model, 
$M_{TA}$: Size of input to TA model, 
$N_{S}$: Size of output of Student model, 
$N_{TA}$: Size of output of TA model \\ 
$I_{S} \in \mathcal{R}^{M_S}$: Input for Student model, 
$I_{TA} \in \mathcal{R}^{M_{TA}}$: Input for TA model 

\Ensure \\
$O_{S} \in \mathcal{R}^{N_{S}}$: Output for Student model, 
$O_{TA} \in \mathcal{R}^{N_{TA}}$: Output for TA model \\ \\
\textbf{Initialize weights}: \\
$\mathbb{W}_S \in \mathcal{R}^{M_S \times N_S}$ \\

$\mathbb{W}_{TA}^{1} \in \mathcal{R}^{(M_{TA}-M_{S}) \times N_S}$ \\

$\mathbb{W}_{TA}^{2} \in \mathcal{R}^{M_{TA} \times (N_{TA} - N_S)}$ \\

Note that $\mathbb{W}_{TA} := \{\mathbb{W}_S, \mathbb{W}_{TA}^{1}, \mathbb{W}_{TA}^{2}  \} $ \\ \\

\textbf{Procedure}: \\

// Split $I_{TA}$ into two parts: \\
$[I_{TA}^0, I_{TA}^{Extra}] \gets I_{TA}$ \,\,\, s.t. $I_{TA}^0 \in \mathcal{R}^{M_S}$ and $I_{TA}^{Extra} \in \mathcal{R}^{(M_{TA}-M_S)}$ \\ \\

// Output for Student model: \\
$O_S$ =  $I_{S}  \mathbb{W}_S$ \\ \\

// Construct Output for TA model: \\
$O_{TA}^0$ = $I_{TA}^0  \mathbb{W}_S$ \\
$O_{TA}^{Extra}$ = $I_{TA}^{Extra}  \mathbb{W}_{TA}^1$ \\

$O_{TA}^1$ = ADD($[O_{TA}^0, O_{TA}^{Extra}]$) \\
$O_{TA}^{2}$ = $I_{TA}  \mathbb{W}_{TA}^{2}$ \\
$O_{TA}$ = CONCATENATE($[O_{TA}^1, O_{TA}^2]$) \\

\end{algorithmic}
\end{algorithm}

\subsubsection{Depth M-nesting}
\label{sec:depth_expansion}
In Depth M-nesting of a given original model, additional layers are added to it that are exclusive to the TA model. This approach is inspired from~\cite{teerapittayanon2016branchynet} where a neural network is augmented with branch classifiers and can skip a few layers when the model is confident in its prediction. 

\subsubsection{d\_model M-nesting}
\label{sec:d_model_expansion}
Another dimension along which we can increase the capacity of the original model is by increasing the dimensionality of hidden representations used in the model, i.e., $d\_model$. 

\subsection{Second Order Optimizer}

For training our proposed nested architecture, we observed that second-order optimizers, such as Shampoo~\cite{gupta2018shampoo}, offer significant improvements over first-order methods like Adagrad~\cite{duchi2011adaptive} and Adam~\cite{kingma2014adam}.  Unlike first-order methods, which precondition each parameter independently, Shampoo-based algorithms precondition the gradient $G$ using a Kronecker-based preconditioner. This results in a preconditioned gradient of $(L\otimes R) vec(G) = vec(LGR)$, where L and R are symmetric left and right preconditioners and $\otimes$ indicates a Kronecker product.  Shampoo has demonstrated improved performance over first-order methods in tasks like click-through rate prediction~\cite{anil2022factory} and smaller-scale large language model (LLM) training~\cite{shi2023distributed}, and we find that it provides additive or even super-additive gains when combined with the \methodname approach, as shown in Section~\ref{sec:ablation-optimizer}.

\begin{figure}
    \centering
    \includegraphics[width=0.4\textwidth]{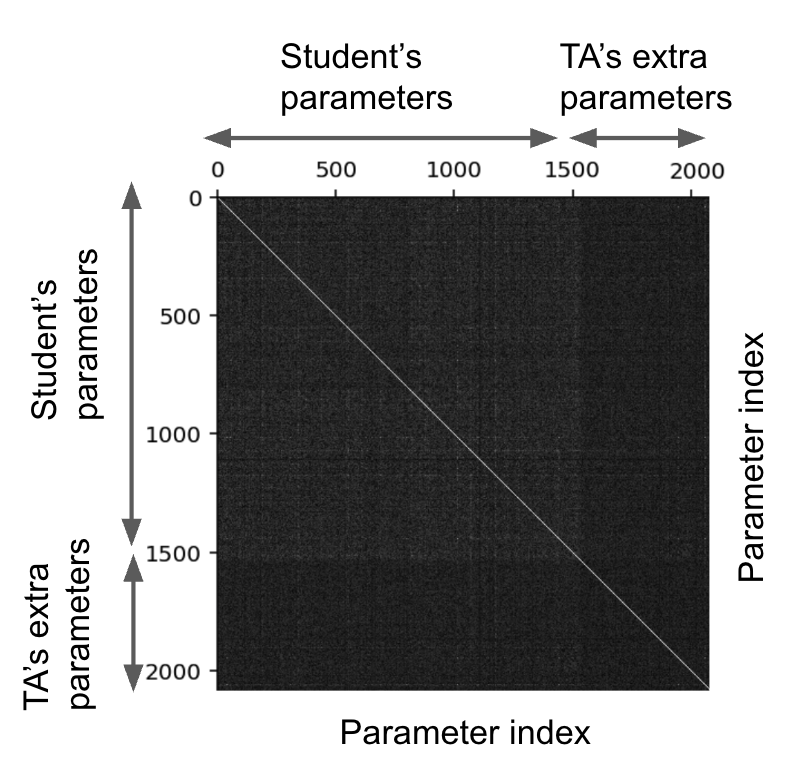}
    \caption{The Shampoo preconditioner matrix for MatTA. We observe that the preconditioner automatically captures parameter correlations within the Student and TA models, which leads to superior performance for training MatTA.}
    \label{fig:preconditioner}
\end{figure}

To understand why second-order optimizers perform well with MatTA, we visualize the preconditioner matrix of a sample run in Figure~\ref{fig:preconditioner}. This matrix corresponds to a nested layer where the Student contains approximately 1,500 neurons and the TA approximately 2,000. Lighter color indicates higher correlation value. As the figure shows, the preconditioner matrix exhibits stronger inter-block correlations within the $1,500\times 1,500$ and $2,000 \times 2,000$ blocks. This suggests that second-order optimizers effectively leverage the parameter correlations within each sub-model, leading to improved training performance on MatTA models.

\section{Experimental Results}\label{sec:exps}

Given the practical utility of MatTA, we demonstrate its effectiveness on a collection of proprietary models and datasets,  both in terms of offline model quality improvements, as well as a large scale online A/B experiment.  We also demonstrate the general applicability of our method on a publicly available serving-sized Transformer model (GPT-2 Medium). Finally, we present results from multiple ablations conducted to understand the importance and interplay of various components in MatTA. 

\subsection{MatTA Deployed in Production ML Models}
Below we discuss two case studies of deploying  MatTA on two production ML models in a large scale recommendation system. These production models use an online training paradigm using the infrastructure detailed in \cite{adbrain2025paper}. 
While we describe the high-level data schema, scale and offline metrics, note that detailed specifications of the two  proprietary datasets, production models (including the architecture and feature-set) and the models' specific use case within the broader system are not disclosed due to their proprietary nature.

\subsubsection{Case Study 1: Relevance Model Data}
\paragraph{\bf Problem and Dataset.} The Relevance Model predicts the relevance of an item for each user. The dataset comprises of features representing the $\{user, context, item\}$ tuple. Ground truth labels for this task come from a very large Teacher model which is trained once, and cannot be used to serve live requests. This dataset has of the order of tens of billions of events that are sampled over the period of a year.

{\bf Results.}
 Table~\ref{tab:production_matta_short} shows the effect of \methodname in terms of model quality improvements. The relative improvement in the evaluation metrics for both the Student model and the TA model are shown in comparison to the production models. The metric of significance is the test AucLoss, which is defined as $1 - AUROC$. Note that the Student model (row 1) is of the same size and architecture as the baseline production model, while the TA model is about 4x larger in selected layers. Naturally, as the TA model is much larger, it performs better than the production baseline but it cannot be deployed due to cost. However, row~1 shows that MatTA's Student model itself is $8.05\%$ more accurate than the production baseline despite being of the {\em same} size and hence is servable with no increase in serving costs. 
 
 {\bf Current Status.} The model has been launched in production following a live A/B experiment which demonstrated a 20\% improvement on a key metric.

\subsubsection{Case Study 2: Quality Model Data}
\paragraph{\bf Problem and Dataset.} The Quality Model is tasked with predicting quality of incoming requests in a massive scale industrial recommendation system. The dataset spans one month and has tens of billions of events. The ground truth comes from user event logs.

{\bf Results.} Similar to the Relevance Model, we measure the quality of the model using test AucLoss, defined as $1 - AUROC$. Here again, row 3 shows MatTA's Student model is 0.6\% more accurate in terms of test AucLoss; note that this model is the same size and architecture as production baseline. Quality improvement due to the TA model is also along the expected lines.

\begin{table}[h!]
\caption{Numbers provided are the \% offline improvements as compared to corresponding production models. AucLoss is defined as $1 - AUROC$. The scales of offline gains seen in the Relevance and Quality models are both internally categorized as a medium to large sized improvement in the model quality. MatTA is launched in production for Relevance model.}
\centering
\begin{tabular}{|p{0.5\linewidth} | c |} 
 \hline
 \textbf{Offline improvements} & \makecell{\textbf{\% Improvement in AucLoss} \\ \textbf{(more negative is better)}}  \\
 
  \hline
MatTA {\bf Student} (Relevance model, Case Study 1) & -8.05\% \\
\hline
MatTA {\bf TA} (Relevance model, Case Study 1) & -9.47\% \\
\hline
\hline
MatTA {\bf Student} (Quality model, Case Study 2) & -0.6\% \\
 \hline
MatTA {\bf TA} (Quality model, Case Study 2) & -0.81\% \\
 \hline
\end{tabular}
\label{tab:production_matta_short}
\end{table}

\subsection{MatTA results on Public Datasets and Models}

In addition to demonstrating the impact of MatTA on proprietary models and dataset, we also apply our approach to a publicly available serving-sized decoder model, GPT-2 Medium \cite{gpt2}. We chose this model since it has approximately 355 million trainable parameters and is of a size that can be easily used in industrial ML systems.

For pre-training the GPT-2 models used in our experiments, we employed the Colossal Clean Crawled Corpus (C4~\cite{C4}).  This corpus is a filtered version of Common Crawl, designed for improved language model training.  C4 undergoes significant cleaning to eliminate low-quality web content, such as boilerplate and navigational elements, yielding a substantial dataset of relatively high-quality text. We train a decoder model with GPT-2 Medium \cite{gpt2} architecture from scratch on 100B tokens from C4 \cite{C4} corpus.

\subsubsection{Evaluation Datasets for GPT-2 Medium}
\label{sec:gpt-datasets}

\hfill \\
For evaluating the model quality of baseline and MatTA GPT-2 models, we employ the six datasets below.

\begin{enumerate}

\item \textbf{LAMBADA.}  Tests the capabilities of language models for text understanding by means of a word prediction task \cite{lambada-dataset}. This dataset focuses on discourse-level coherence.

\item \textbf{ARC Easy.} From AI2 Reasoning Challenge, this dataset evaluates basic scientific reasoning with multiple-choice questions designed for elementary school level science \cite{arc-dataset}.

\item \textbf{ARC Challenging.} From AI2 Reasoning Challenge, this dataset assesses more advanced scientific reasoning and knowledge with complex, multiple-choice questions requiring deeper inference \cite{arc-dataset}. This dataset is significantly harder than ARC Easy.

\item \textbf{SAT Math.}  This dataset measures mathematical reasoning ability using SAT-style math problems \cite{sat-agieval}. Given our focus on small model size, we use ``Inclusion accuracy'' as the metric, which scores 1 for an example if the correct answer is included in the model output.

\item \textbf{HellaSwag.} This dataset evaluates commonsense reasoning through completion \cite{hellaswag}. Models must select the most plausible continuation from adversarially generated, deceptively plausible but incorrect options.

\item \textbf{BoolQ.}  This dataset tests reading comprehension and boolean (yes/no) question answering \cite{boolq}. Models are given a passage and a question, and must answer ``yes'' or ``no'' based on the passage content.

\end{enumerate}

\subsubsection{Offline model improvements in GPT-2 Medium}
\hfill \\
To demonstrate the effectiveness of \methodname on a servable sized public model, we use the GPT-2 Medium model as the original Student model. To create the corresponding TA model, we employ Layer \textit{M-nesting} to expand the hidden dimension in FFN module to twice its original size in every Transformer block (Section~\ref{sec:layer_expansion}). In addition, we also employ Depth \textit{M-nesting} (Section~\ref{sec:depth_expansion}) and add 12 more layers, amounting to a total of 36 layers in the TA model. We then co-train the Student and the TA models using \methodname on C4 dataset for 100 billion tokens.

Table \ref{tab:gpt2_eval} presents the performance of our proposed \methodname models alongside baseline GPT-2 Medium and a larger GPT-3 Medium model (from Appendix in \cite{GPT3_neurips}, used as an upper bound reference here) across a diverse set of benchmark tasks designed to evaluate various aspects of language understanding and reasoning. Evaluations on the above datasets are performed zero-shot without any finetuning or in-context examples. As shown, both the \methodname GPT-2 Student and the corresponding TA model demonstrate consistent improvements over baseline GPT-2 Medium across all tasks, including long-range context understanding (LAMBADA), scientific reasoning (ARC Easy \& Challenging), mathematical reasoning (SAT Math), commonsense reasoning (HellaSwag), and question answering (BoolQ). While the GPT-3 Medium model, as expected due to being next generation and being trained on much more data, exhibits superior performance overall, the significant improvements observed by \methodname GPT-2 and the TA model – both derived from original GPT-2 Medium architecture – underscore the efficacy of MatTA in efficiently generating higher quality models from an existing serving model.

\begin{table*}[t]
\caption{\methodname helps GPT2 improve in all the popular benchmarks we evaluated on. The quality of the MatTA Student model is, as expected, less than the quality of the TA model since the former sub-model distills from the TA. } 
\centering
\begin{tabular}{ |l|c|c|c|c|c|c| }
\hline
  \textbf{Zero-shot accuracy from Intervention} & \textbf{LAMBADA} & \textbf{ARC} & \textbf{ARC} & \textbf{SAT Math} & \textbf{HellaSwag} & \textbf{BoolQ} \\
 &  & \textbf{easy} & \textbf{challenging} &  &  &  \\
\hline
GPT2 & 27.56 & 42.84 & 23.72 & 31.81 & 31.61 & 59.51 \\
\hline
\methodname GPT2 (without Shampoo) & 30.5 & 43.27 & 24.15 & 39.54 & 32.24 & 60.00 \\
\hline
\methodname GPT2 (with Shampoo) & 32.3 & 42.34 & 22.78 & 53.63 & 32.08 & 61.34 \\
\hline
TA Model & 31.73 & 43.73 & 23.46 & 60.45 & 34.87 & 60.00 \\
\hline
GPT-3 Medium (for reference) & 54.3 & 46.5 & 29.5 & N/A & 43.6 & 60.3 \\
\hline
\end{tabular}
\label{tab:gpt2_eval} 
\end{table*}

\subsubsection{Quality of extracted sub-models}

\begin{figure}[h]
    \centering
    \includegraphics[width=0.45\textwidth]{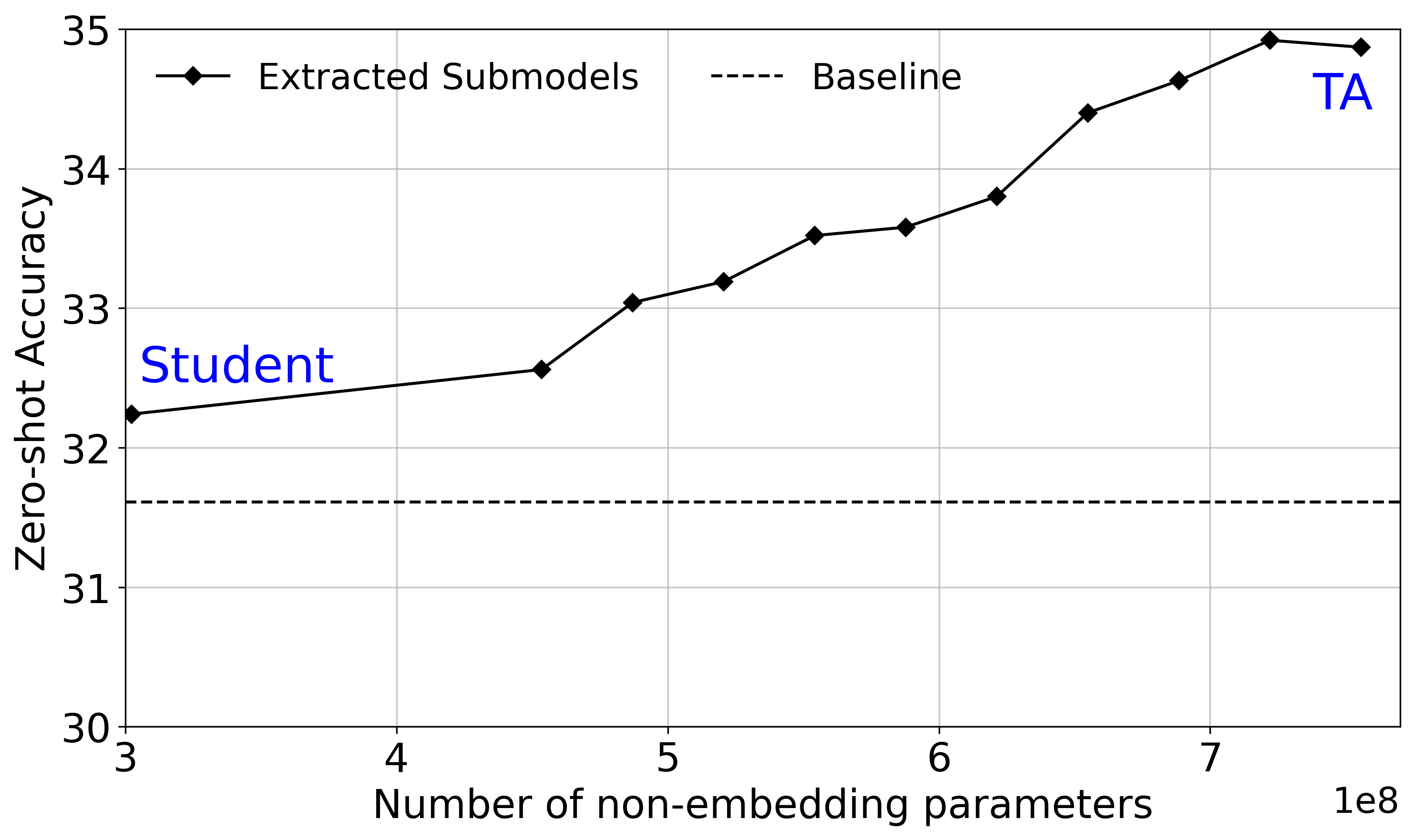}
    \caption{Performance of extracted sub-models from MatTA GPT2-Medium on HellaSwag \newline}
    \label{fig:mixm}
\end{figure}

\hfill \\

We extract a range of Student models from MatTA GPT-2 Medium model using Mix'n'Match~\cite{devvrit2024matformer}, and evaluate the extracted checkpoints on public evaluation benchmark (Section~\ref{sec:gpt-datasets}). Given that GPT2-Medium is a stack of identical Transformer layers, the number of sub-models that can be extracted from the trained MatTA TA can be large. We follow a \textit{wide-narrow-wide} procedure that empirically allows extraction of high quality sub-models. In this procedure, we use the lowest $K/2$ layers and the uppermost $K/2$ layers from the MatTA TA model. The remaining layers are extracted from the MatTA Student. 

Our baseline GPT-2 Medium, and MatTA Student model both have 24 Transformer layers and FFN hidden size of 4096. The MatTA TA model has 36 Transformer layers and FFN hidden size of 8192. We start with a full narrow model with 36 layers ($K=0$) and progressively increase $K$. As a result, the $(N-K)$ layers comprising middle section of the extracted student model are "narrow", while the other $K$ are "wide". We do this for $K=\{0, 4, 8, 12, 16, 20, 24, 28, 32, 36\}$. This procedure allowed us to generate Student sub-models with increasing sizes, effectively exploring the trade-off between model size and quality. Figure \ref{fig:mixm} presents the zero-shot accuracy achieved by the extracted Student sub-models across a spectrum of non-embedding parameter counts. As depicted, we observe a clear trend of increasing zero-shot accuracy as the number of non-embedding parameters grows. All extracted models from MatTA in Figure \ref{fig:mixm} are of a stronger quality than the baseline GPT-2 Medium model without MatTA. This demonstrates the ability of our method to generate a range of servable Student models with compelling performance characteristics at different size constraints.

\subsection{Ablations}

\subsubsection{Effect of second-order optimizer Shampoo in MatTA}
\label{sec:ablation-optimizer}
\hfill \\
In this section, we present offline experimental results  for model quality improvements from \methodname and the confounding effect of its different components. We start with the impact of second-order optimizer Shampoo \cite{shampoo} on the Matryoshka architecture imposed by MatTA.

\begin{table}[h!]
\caption{Ablations to understand offline model improvements from \methodname and second-order Shampoo optimizer (\% is calculated vs baseline). For Relevance and Quality models, a reduction of 3.0\% and 0.25\% respectively are considered significant improvements, leading to consequential improvements in key metrics. While \methodname and Shampoo individually improve model quality of Student, their combined effect is super-additive.}
\centering
\begin{tabular}{|l | cc | cc |} 
 \hline
 \textbf{Intervention} & \multicolumn{2}{c|}{\textbf{Relevance Model}} & \multicolumn{2}{c|}{\textbf{Quality Model}}  \\
 
  & Student & TA & Student & TA \\
  \hline
Shampoo only & -3.89\% & NA & -0.28\% & NA \\ 
 \hline 
\methodname without Shampoo & -4.63\% & -5.92\% & -0.25\% & -0.32\%  \\ 
 \hline
\methodname with Shampoo & -8.05\% & -9.47\% & -0.6\% & -0.81\%  \\ 
 \hline
\end{tabular}
\label{tab:original_improvements}
\end{table}

For Relevance and Quality models, a reduction of 3.0\% and 0.25\% respectively are considered a significant improvement.
The baseline production models are trained with first order optimizers such as Adam~\cite{adam}, Adagrad~\cite{duchi2011adaptive} or SM3~\cite{sm3}.

Table~\ref{tab:original_improvements} shows the ablation of the nested co-training approach of MatTA with choice of second-order optimizer. We note that changing the optimizer alone from first order to second order (Shampoo~\cite{shampoo}) brings significant model improvement. Typically, the incremental benefit of successive model improvements typically diminishes, with the combined gain often substantially less than the sum of individual contributions. However in the case of \methodname and Shampoo, we observe \textit{super-additive} improvements. This means that, as seen in Table~\ref{tab:original_improvements}, the final improvement is generally close to or more than the sum of individual improvements due to \methodname and the optimizer. This shows that the second order optimizer Shampoo is especially suitable at exploiting the nested structure of parameters in MatTA. Using Shampoo optimizer requires approximately 15-25\% more training compute and time than first-order methods because it involves approximating preconditioners derived from second-order information.

\subsubsection{Effect of Parameter sharing in MatTA}
\hfill \\
Parameter sharing has been shown to improve model efficiency and overall model sizes~\cite{ullrich2017soft}. In addition, as shown in works focused on Recommendation systems~\cite{meng2020incorporating}, Speech recognition~\cite{yu2020dual} and in Computer vision, parameter sharing might also help achieve better representations and improved model quality. At the same time however, parameter sharing imposes a constraint on the Student and the \methodname models by tying their weights with each other. This may prevent the Student and the TA models both from independently updating their parameters, restricting their degrees of freedom. 
Thus, we articulate the following competing hypotheses with respect to parameter sharing in MatTA:

\begin{itemize}
    \item H1: Parameter sharing helps generalize extracted Students better and improves their quality.
    \item H2: Parameter sharing restricts overall capacity of MatTA and hurts quality of extracted models.
\end{itemize}

Note that parameter sharing is an essential component of \methodname since it enables extraction of a range of models of varying sizes and qualities from the trained model. The research question that we would like to answer is: what is the effect of parameter sharing in \methodname on Student and TA \methodname models? In other words, we would like to understand which of the above two hypotheses (H1 and H2) dominates empirically.

\begin{table}[h!]
\caption{Metric improvements for \methodname Student model with and without parameter sharing. Due to limitations of the training framework, the embedding table is shared between Student and TA \methodname models. Note that the \% numbers differ from Table~\ref{tab:original_improvements} due to difference in underlying dates that the models are trained over.}
\centering
\begin{tabular}{|m{9em} | cc | cc|} 
 \hline
 \textbf{Loss improvement vs baseline \%} & \multicolumn{2}{c|}{\textbf{Relevance Model}} & \multicolumn{2}{c|}{\textbf{Quality Model}}  \\
 & Student & TA & Student & TA \\
 \hline
\methodname (with parameter sharing) & -8.03\% & -8.76\% & -0.53\% &  -0.74\%  \\ 
 \hline 
Without parameter sharing & -8.27\% & -9.16\% &  -0.67\% & -0.99\% \\ 
 \hline
\end{tabular}
\label{tab:weight_sharing}
\end{table}

 As Table~\ref{tab:weight_sharing} shows, parameter sharing leads to a slight degradation in the model quality of both the Student and the TA models when compared to training without parameter sharing. This suggests that the second hypothesis (H2) ends up dominating over the first as outlined earlier in this section. An important takeaway from Table~\ref{tab:weight_sharing} is that if the goal is primarily to improve the serving Student model, then having completely independent parameters of the Student and TA is recommended. Parameter sharing, however, is necessary in order to extract a range of models from MatTA.

\section{Conclusion}
In this paper, we propose an easy-to-use elastic distillation and co-training framework called Matryoshka Teaching Assistant (MatTA) which addresses the need for elastic serving Student models, and for more relatable models to distill from.  We do so by first \textit{M-nesting} a serving model to create a Teaching Assistant (TA) model, and then training the Student and the TA using online distillation. The value of MatTA is demonstrated by extraction of a spectrum of models offering a compelling trade-off between quality and size. At the same time, MatTA can also lead to significant quality gains in baseline student models. We demonstrate the efficacy of our approach on proprietary models and on a publicly available model: GPT-2 Medium. Through experiments studying confounding effects of various components of the MatTA approach, we showcase the super-additive nature of second-order optimizers with Matryoshka model structure. The proposed Teacher-TA-Student recipe is widely applicable since compared to existing approaches, it does not necessitate having a super-network or an ensemble of models to extract subs-models from. MatTA has been launched into production and has led to an improvement of over 20\% on a key metric.


\bibliographystyle{ACM-Reference-Format}
\bibliography{references}


\begin{thebibliography}{55}


\ifx \showCODEN    \undefined \def \showCODEN     #1{\unskip}     \fi
\ifx \showISBNx    \undefined \def \showISBNx     #1{\unskip}     \fi
\ifx \showISBNxiii \undefined \def \showISBNxiii  #1{\unskip}     \fi
\ifx \showISSN     \undefined \def \showISSN      #1{\unskip}     \fi
\ifx \showLCCN     \undefined \def \showLCCN      #1{\unskip}     \fi
\ifx \shownote     \undefined \def \shownote      #1{#1}          \fi
\ifx \showarticletitle \undefined \def \showarticletitle #1{#1}   \fi
\ifx \showURL      \undefined \def \showURL       {\relax}        \fi
\providecommand\bibfield[2]{#2}
\providecommand\bibinfo[2]{#2}
\providecommand\natexlab[1]{#1}
\providecommand\showeprint[2][]{arXiv:#2}

\bibitem[Anil et~al\mbox{.}(2022)]%
        {anil2022factory}
\bibfield{author}{\bibinfo{person}{Rohan Anil}, \bibinfo{person}{Sandra
  Gadanho}, \bibinfo{person}{Da Huang}, \bibinfo{person}{Nijith Jacob},
  \bibinfo{person}{Zhuoshu Li}, \bibinfo{person}{Dong Lin},
  \bibinfo{person}{Todd Phillips}, \bibinfo{person}{Cristina Pop},
  \bibinfo{person}{Kevin Regan}, \bibinfo{person}{Gil~I Shamir},
  {et~al\mbox{.}}} \bibinfo{year}{2022}\natexlab{}.
\newblock \showarticletitle{On the factory floor: {ML} engineering for
  industrial-scale ads recommendation models}.
\newblock \bibinfo{journal}{\emph{arXiv preprint arXiv:2209.05310}}
  (\bibinfo{year}{2022}).
\newblock


\bibitem[Anil et~al\mbox{.}(2020)]%
        {anil2020scalable}
\bibfield{author}{\bibinfo{person}{Rohan Anil}, \bibinfo{person}{Vineet Gupta},
  \bibinfo{person}{Tomer Koren}, \bibinfo{person}{Kevin Regan}, {and}
  \bibinfo{person}{Yoram Singer}.} \bibinfo{year}{2020}\natexlab{}.
\newblock \showarticletitle{Scalable second order optimization for deep
  learning}.
\newblock \bibinfo{journal}{\emph{arXiv preprint arXiv:2002.09018}}
  (\bibinfo{year}{2020}).
\newblock


\bibitem[Anil et~al\mbox{.}(2019)]%
        {sm3}
\bibfield{author}{\bibinfo{person}{Rohan Anil}, \bibinfo{person}{Vineet Gupta},
  \bibinfo{person}{Tomer Koren}, {and} \bibinfo{person}{Yoram Singer}.}
  \bibinfo{year}{2019}\natexlab{}.
\newblock \showarticletitle{Memory efficient adaptive optimization}.
\newblock \bibinfo{journal}{\emph{Advances in Neural Information Processing
  Systems}}  \bibinfo{volume}{32} (\bibinfo{year}{2019}).
\newblock


\bibitem[Anil et~al\mbox{.}(2018)]%
        {onlinedistillation}
\bibfield{author}{\bibinfo{person}{Rohan Anil}, \bibinfo{person}{Gabriel
  Pereyra}, \bibinfo{person}{Alexandre Passos}, \bibinfo{person}{Robert
  Ormandi}, \bibinfo{person}{George~E Dahl}, {and} \bibinfo{person}{Geoffrey~E
  Hinton}.} \bibinfo{year}{2018}\natexlab{}.
\newblock \showarticletitle{Large scale distributed neural network training
  through online distillation}.
\newblock \bibinfo{journal}{\emph{arXiv preprint arXiv:1804.03235}}
  (\bibinfo{year}{2018}).
\newblock


\bibitem[Brown(2020)]%
        {GPT3_neurips}
\bibfield{author}{\bibinfo{person}{Benjamin Mann Nick Ryder Melanie Subbiah
  Jared D. Kaplan Prafulla Dhariwal Arvind Neelakantan et~al. Brown, Tom}.}
  \bibinfo{year}{2020}\natexlab{}.
\newblock \showarticletitle{Language models are few-shot learners}.
\newblock \bibinfo{journal}{\emph{Advances in Neural Information Processing
  Systems}}  \bibinfo{volume}{33}, \bibinfo{pages}{1877--1901}.
\newblock


\bibitem[Cai et~al\mbox{.}(2019)]%
        {cai2019once}
\bibfield{author}{\bibinfo{person}{Han Cai}, \bibinfo{person}{Chuang Gan},
  \bibinfo{person}{Tianzhe Wang}, \bibinfo{person}{Zhekai Zhang}, {and}
  \bibinfo{person}{Song Han}.} \bibinfo{year}{2019}\natexlab{}.
\newblock \showarticletitle{Once-for-all: Train one network and specialize it
  for efficient deployment}.
\newblock \bibinfo{journal}{\emph{arXiv preprint arXiv:1908.09791}}
  (\bibinfo{year}{2019}).
\newblock


\bibitem[Cai et~al\mbox{.}(2024)]%
        {cai2024flextron}
\bibfield{author}{\bibinfo{person}{Ruisi Cai}, \bibinfo{person}{Saurav
  Muralidharan}, \bibinfo{person}{Greg Heinrich}, \bibinfo{person}{Hongxu Yin},
  \bibinfo{person}{Zhangyang Wang}, \bibinfo{person}{Jan Kautz}, {and}
  \bibinfo{person}{Pavlo Molchanov}.} \bibinfo{year}{2024}\natexlab{}.
\newblock \showarticletitle{Flextron: Many-in-One Flexible Large Language
  Model}.
\newblock \bibinfo{journal}{\emph{arXiv preprint arXiv:2406.10260}}
  (\bibinfo{year}{2024}).
\newblock


\bibitem[Clark(2019)]%
        {boolq}
\bibfield{author}{\bibinfo{person}{Christopher et~al Clark}.}
  \bibinfo{year}{2019}\natexlab{}.
\newblock \showarticletitle{{B}ool{Q}: Exploring the Surprising Difficulty of
  Natural Yes/No Questions}. In \bibinfo{booktitle}{\emph{Proceedings of the
  2019 Conference of the North {A}merican Chapter of the Association for
  Computational Linguistics: Human Language Technologies, Volume 1 (Long and
  Short Papers)}}.
\newblock


\bibitem[Clark et~al\mbox{.}(2018)]%
        {arc-dataset}
\bibfield{author}{\bibinfo{person}{Peter Clark}, \bibinfo{person}{Isaac
  Cowhey}, \bibinfo{person}{Oren Etzioni}, \bibinfo{person}{Tushar Khot},
  \bibinfo{person}{Ashish Sabharwal}, \bibinfo{person}{Carissa Schoenick},
  {and} \bibinfo{person}{Oyvind Tafjord}.} \bibinfo{year}{2018}\natexlab{}.
\newblock \showarticletitle{Think you have solved question answering? try arc,
  the ai2 reasoning challenge}.
\newblock \bibinfo{journal}{\emph{arXiv preprint arXiv:1803.05457}}
  (\bibinfo{year}{2018}).
\newblock


\bibitem[Dehghani et~al\mbox{.}(2023)]%
        {dehghani2023scaling}
\bibfield{author}{\bibinfo{person}{Mostafa Dehghani}, \bibinfo{person}{Josip
  Djolonga}, \bibinfo{person}{Basil Mustafa}, \bibinfo{person}{Piotr
  Padlewski}, \bibinfo{person}{Jonathan Heek}, \bibinfo{person}{Justin Gilmer},
  \bibinfo{person}{Andreas~Peter Steiner}, \bibinfo{person}{Mathilde Caron},
  \bibinfo{person}{Robert Geirhos}, \bibinfo{person}{Ibrahim Alabdulmohsin},
  {et~al\mbox{.}}} \bibinfo{year}{2023}\natexlab{}.
\newblock \showarticletitle{Scaling vision transformers to 22 billion
  parameters}. In \bibinfo{booktitle}{\emph{International Conference on Machine
  Learning}}. PMLR, \bibinfo{pages}{7480--7512}.
\newblock


\bibitem[Devvrit et~al\mbox{.}(2024)]%
        {devvrit2024matformer}
\bibfield{author}{\bibinfo{person}{Fnu Devvrit}, \bibinfo{person}{Sneha
  Kudugunta}, \bibinfo{person}{Aditya Kusupati}, \bibinfo{person}{Tim
  Dettmers}, \bibinfo{person}{Kaifeng Chen}, \bibinfo{person}{Inderjit~S
  Dhillon}, \bibinfo{person}{Yulia Tsvetkov}, \bibinfo{person}{Hannaneh
  Hajishirzi}, \bibinfo{person}{Sham~M. Kakade}, \bibinfo{person}{Ali Farhadi},
  {and} \bibinfo{person}{Prateek Jain}.} \bibinfo{year}{2024}\natexlab{}.
\newblock \showarticletitle{MatFormer: Nested Transformer for Elastic
  Inference}. In \bibinfo{booktitle}{\emph{The Thirty-eighth Annual Conference
  on Neural Information Processing Systems}}.
\newblock


\bibitem[Kusupati et~al\mbox{.}(2022)]%
        {mrl-2022}
\bibfield{author}{\bibinfo{person}{Aditya Kusupati}, \bibinfo{person}{Gantavya
  Bhatt}, \bibinfo{person}{Aniket Rege}, \bibinfo{person}{Matthew
  Wallingford}, \bibinfo{person}{Aditya Sinha}, \bibinfo{person}{Vivek
  Ramanujan}, \bibinfo{person}{William Howard-Snyder}, \bibinfo{person}{Kaifeng
  Chen}, \bibinfo{person}{Sham Kakade}, \bibinfo{person}{Prateek Jain},
  {and} \bibinfo{person}{Ali Farhadi}.} \bibinfo{year}{2022}\natexlab{}.
\newblock \showarticletitle{Matryoshka representation learning}. In \bibinfo{booktitle}{\emph{The Thirty-sixth Annual Conference
  on Neural Information Processing Systems}}.
\newblock


\bibitem[Duchi et~al\mbox{.}(2011a)]%
        {duchi2011adaptive}
\bibfield{author}{\bibinfo{person}{John Duchi}, \bibinfo{person}{Elad Hazan},
  {and} \bibinfo{person}{Yoram Singer}.} \bibinfo{year}{2011}\natexlab{a}.
\newblock \showarticletitle{Adaptive subgradient methods for online learning
  and stochastic optimization.}
\newblock \bibinfo{journal}{\emph{Journal of machine learning research}}
  \bibinfo{volume}{12}, \bibinfo{number}{7} (\bibinfo{year}{2011}).
\newblock


\bibitem[Duchi et~al\mbox{.}(2011b)]%
        {adagrad}
\bibfield{author}{\bibinfo{person}{John Duchi}, \bibinfo{person}{Elad Hazan},
  {and} \bibinfo{person}{Yoram Singer}.} \bibinfo{year}{2011}\natexlab{b}.
\newblock \showarticletitle{Adaptive subgradient methods for online learning
  and stochastic optimization.}
\newblock \bibinfo{journal}{\emph{Journal of machine learning research}}
  \bibinfo{volume}{12}, \bibinfo{number}{7} (\bibinfo{year}{2011}).
\newblock


\bibitem[Duvvuri et~al\mbox{.}({[n.\,d.]})]%
        {duvvuricombining}
\bibfield{author}{\bibinfo{person}{Sai~Surya Duvvuri}, \bibinfo{person}{Fnu
  Devvrit}, \bibinfo{person}{Rohan Anil}, \bibinfo{person}{Cho-Jui Hsieh},
  {and} \bibinfo{person}{Inderjit~S Dhillon}.}
  \bibinfo{year}{[n.\,d.]}\natexlab{}.
\newblock \showarticletitle{{CASPR}: Combining Axes Preconditioners through
  Kronecker Approximation for Deep Learning}. In \bibinfo{booktitle}{\emph{The
  Twelfth International Conference on Learning Representations}}.
\newblock


\bibitem[Fedus et~al\mbox{.}(2022)]%
        {fedus2022switch}
\bibfield{author}{\bibinfo{person}{William Fedus}, \bibinfo{person}{Barret
  Zoph}, {and} \bibinfo{person}{Noam Shazeer}.}
  \bibinfo{year}{2022}\natexlab{}.
\newblock \showarticletitle{Switch transformers: Scaling to trillion parameter
  models with simple and efficient sparsity}.
\newblock \bibinfo{journal}{\emph{Journal of Machine Learning Research}}
  \bibinfo{volume}{23}, \bibinfo{number}{120} (\bibinfo{year}{2022}),
  \bibinfo{pages}{1--39}.
\newblock


\bibitem[Grattafiori et~al\mbox{.}(2024)]%
        {llama2024}
\bibfield{author}{\bibinfo{person}{Aaron Grattafiori},
  \bibinfo{person}{Abhimanyu Dubey}, \bibinfo{person}{Abhinav Jauhri},
  \bibinfo{person}{Abhinav Pandey}, \bibinfo{person}{Abhishek Kadian},
  \bibinfo{person}{Ahmad Al-Dahle}, \bibinfo{person}{Aiesha Letman},
  {et~al\mbox{.}}} \bibinfo{year}{2024}\natexlab{}.
\newblock \showarticletitle{The Llama 3 herd of models}.
\newblock \bibinfo{journal}{\emph{arXiv preprint arXiv:2407.21783}}.
\newblock


\bibitem[Gupta et~al\mbox{.}(2018a)]%
        {shampoo}
\bibfield{author}{\bibinfo{person}{Vineet Gupta}, \bibinfo{person}{Tomer
  Koren}, {and} \bibinfo{person}{Yoram Singer}.}
  \bibinfo{year}{2018}\natexlab{a}.
\newblock \showarticletitle{Shampoo: Preconditioned stochastic tensor
  optimization}. In \bibinfo{booktitle}{\emph{International Conference on
  Machine Learning}}. PMLR, \bibinfo{pages}{1842--1850}.
\newblock


\bibitem[Gupta et~al\mbox{.}(2018b)]%
        {gupta2018shampoo}
\bibfield{author}{\bibinfo{person}{Vineet Gupta}, \bibinfo{person}{Tomer
  Koren}, {and} \bibinfo{person}{Yoram Singer}.}
  \bibinfo{year}{2018}\natexlab{b}.
\newblock \showarticletitle{Shampoo: Preconditioned stochastic tensor
  optimization}. In \bibinfo{booktitle}{\emph{International Conference on
  Machine Learning}}. PMLR, \bibinfo{pages}{1842--1850}.
\newblock


\bibitem[Hinton et~al\mbox{.}(2015)]%
        {distillation}
\bibfield{author}{\bibinfo{person}{Geoffrey Hinton}, \bibinfo{person}{Oriol
  Vinyals}, {and} \bibinfo{person}{Jeff Dean}.}
  \bibinfo{year}{2015}\natexlab{}.
\newblock \showarticletitle{Distilling the knowledge in a neural network}.
\newblock \bibinfo{journal}{\emph{CoRR}}  \bibinfo{volume}{abs/1503.02531}
  (\bibinfo{year}{2015}).
\newblock


\bibitem[Hua et~al\mbox{.}(2022)]%
        {gauunit}
\bibfield{author}{\bibinfo{person}{Weizhe Hua}, \bibinfo{person}{Zihang Dai},
  \bibinfo{person}{Hanxiao Liu}, {and} \bibinfo{person}{Quoc Le}.}
  \bibinfo{year}{2022}\natexlab{}.
\newblock \showarticletitle{Transformer quality in linear time}. In
  \bibinfo{booktitle}{\emph{International conference on machine learning}}.
  PMLR, \bibinfo{pages}{9099--9117}.
\newblock


\bibitem[Kingma(2014a)]%
        {kingma2014adam}
\bibfield{author}{\bibinfo{person}{Diederik~P Kingma}.}
  \bibinfo{year}{2014}\natexlab{a}.
\newblock \showarticletitle{Adam: A method for stochastic optimization}.
\newblock \bibinfo{journal}{\emph{arXiv preprint arXiv:1412.6980}}
  (\bibinfo{year}{2014}).
\newblock


\bibitem[Kingma(2014b)]%
        {adam}
\bibfield{author}{\bibinfo{person}{Diederik~P Kingma}.}
  \bibinfo{year}{2014}\natexlab{b}.
\newblock \showarticletitle{Adam: A method for stochastic optimization}.
\newblock \bibinfo{journal}{\emph{arXiv preprint arXiv:1412.6980}}
  (\bibinfo{year}{2014}).
\newblock


\bibitem[Kurata and Saon(2020)]%
        {kurata2020knowledge}
\bibfield{author}{\bibinfo{person}{Gakuto Kurata} {and} \bibinfo{person}{George
  Saon}.} \bibinfo{year}{2020}\natexlab{}.
\newblock \showarticletitle{Knowledge Distillation from Offline to Streaming
  RNN Transducer for End-to-End Speech Recognition.}. In
  \bibinfo{booktitle}{\emph{Interspeech}}. \bibinfo{pages}{2117--2121}.
\newblock


\bibitem[Kurian et~al\mbox{.}(2025)]%
        {adbrain2025paper}
\bibfield{author}{\bibinfo{person}{George Kurian}, \bibinfo{person}{Somayeh
  Sardashti}, \bibinfo{person}{Ryan Sims}, \bibinfo{person}{Felix Berger},
  \bibinfo{person}{Gary Holt}, \bibinfo{person}{Yang Li},
  \bibinfo{person}{Jeremiah Willcock}, \bibinfo{person}{Kaiyuan Wang},
  \bibinfo{person}{Herve Quiroz}, \bibinfo{person}{Abdulrahman Salem}, {and}
  \bibinfo{person}{Julian Grady}.} \bibinfo{year}{2025}\natexlab{}.
\newblock \bibinfo{title}{Scalable Machine Learning Training Infrastructure for
  Online Ads Recommendation and Auction Scoring Modeling at Google}.
\newblock
\showeprint[arxiv]{2501.10546}~[cs.DC]
\urldef\tempurl%
\url{https://arxiv.org/abs/2501.10546}
\showURL{%
\tempurl}


\bibitem[Lan et~al\mbox{.}(2019)]%
        {lan2019self}
\bibfield{author}{\bibinfo{person}{Xu Lan}, \bibinfo{person}{Xiatian Zhu},
  {and} \bibinfo{person}{Shaogang Gong}.} \bibinfo{year}{2019}\natexlab{}.
\newblock \showarticletitle{Self-referenced deep learning}. In
  \bibinfo{booktitle}{\emph{Computer Vision--ACCV 2018: 14th Asian Conference
  on Computer Vision, Perth, Australia, December 2--6, 2018, Revised Selected
  Papers, Part II 14}}. Springer, \bibinfo{pages}{284--300}.
\newblock


\bibitem[Li et~al\mbox{.}(2022)]%
        {li2022blip}
\bibfield{author}{\bibinfo{person}{Junnan Li}, \bibinfo{person}{Dongxu Li},
  \bibinfo{person}{Caiming Xiong}, {and} \bibinfo{person}{Steven Hoi}.}
  \bibinfo{year}{2022}\natexlab{}.
\newblock \showarticletitle{Blip: Bootstrapping language-image pre-training for
  unified vision-language understanding and generation}. In
  \bibinfo{booktitle}{\emph{International conference on machine learning}}.
  PMLR, \bibinfo{pages}{12888--12900}.
\newblock


\bibitem[Li et~al\mbox{.}(2021)]%
        {li2021align}
\bibfield{author}{\bibinfo{person}{Junnan Li}, \bibinfo{person}{Ramprasaath
  Selvaraju}, \bibinfo{person}{Akhilesh Gotmare}, \bibinfo{person}{Shafiq
  Joty}, \bibinfo{person}{Caiming Xiong}, {and} \bibinfo{person}{Steven
  Chu~Hong Hoi}.} \bibinfo{year}{2021}\natexlab{}.
\newblock \showarticletitle{Align before fuse: Vision and language
  representation learning with momentum distillation}.
\newblock \bibinfo{journal}{\emph{Advances in neural information processing
  systems}}  \bibinfo{volume}{34} (\bibinfo{year}{2021}),
  \bibinfo{pages}{9694--9705}.
\newblock


\bibitem[Li et~al\mbox{.}(2023)]%
        {li2023prompt}
\bibfield{author}{\bibinfo{person}{Lei Li}, \bibinfo{person}{Yongfeng Zhang},
  {and} \bibinfo{person}{Li Chen}.} \bibinfo{year}{2023}\natexlab{}.
\newblock \showarticletitle{Prompt distillation for efficient llm-based
  recommendation}. In \bibinfo{booktitle}{\emph{Proceedings of the 32nd ACM
  International Conference on Information and Knowledge Management}}.
  \bibinfo{pages}{1348--1357}.
\newblock


\bibitem[Liang et~al\mbox{.}(2023)]%
        {pmlr-v202-liang23j}
\bibfield{author}{\bibinfo{person}{Chen Liang}, \bibinfo{person}{Simiao Zuo},
  \bibinfo{person}{Qingru Zhang}, \bibinfo{person}{Pengcheng He},
  \bibinfo{person}{Weizhu Chen}, {and} \bibinfo{person}{Tuo Zhao}.}
  \bibinfo{year}{2023}\natexlab{}.
\newblock \showarticletitle{Less is More: Task-aware Layer-wise Distillation
  for Language Model Compression}. In \bibinfo{booktitle}{\emph{Proceedings of
  the 40th International Conference on Machine Learning}}
  \emph{(\bibinfo{series}{Proceedings of Machine Learning Research},
  Vol.~\bibinfo{volume}{202})}, \bibfield{editor}{\bibinfo{person}{Andreas
  Krause}, \bibinfo{person}{Emma Brunskill}, \bibinfo{person}{Kyunghyun Cho},
  \bibinfo{person}{Barbara Engelhardt}, \bibinfo{person}{Sivan Sabato}, {and}
  \bibinfo{person}{Jonathan Scarlett}} (Eds.). \bibinfo{publisher}{PMLR},
  \bibinfo{pages}{20852--20867}.
\newblock


\bibitem[Liu et~al\mbox{.}(2019)]%
        {liu2019end}
\bibfield{author}{\bibinfo{person}{Yuchen Liu}, \bibinfo{person}{Hao Xiong},
  \bibinfo{person}{Zhongjun He}, \bibinfo{person}{Jiajun Zhang},
  \bibinfo{person}{Hua Wu}, \bibinfo{person}{Haifeng Wang}, {and}
  \bibinfo{person}{Chengqing Zong}.} \bibinfo{year}{2019}\natexlab{}.
\newblock \showarticletitle{End-to-end speech translation with knowledge
  distillation}.
\newblock \bibinfo{journal}{\emph{arXiv preprint arXiv:1904.08075}}
  (\bibinfo{year}{2019}).
\newblock


\bibitem[Markov and Matsui(2016)]%
        {markov2016robust}
\bibfield{author}{\bibinfo{person}{Konstantin Markov} {and}
  \bibinfo{person}{Tomoko Matsui}.} \bibinfo{year}{2016}\natexlab{}.
\newblock \showarticletitle{Robust speech recognition using generalized
  distillation framework.}. In \bibinfo{booktitle}{\emph{Interspeech}}.
  \bibinfo{pages}{2364--2368}.
\newblock


\bibitem[Martens and Grosse(2015)]%
        {martens2015optimizing}
\bibfield{author}{\bibinfo{person}{James Martens} {and} \bibinfo{person}{Roger
  Grosse}.} \bibinfo{year}{2015}\natexlab{}.
\newblock \showarticletitle{Optimizing neural networks with
  {K}ronecker-factored approximate curvature}. In
  \bibinfo{booktitle}{\emph{International conference on machine learning}}.
  PMLR, \bibinfo{pages}{2408--2417}.
\newblock


\bibitem[Meng et~al\mbox{.}(2020)]%
        {meng2020incorporating}
\bibfield{author}{\bibinfo{person}{Wenjing Meng}, \bibinfo{person}{Deqing
  Yang}, {and} \bibinfo{person}{Yanghua Xiao}.}
  \bibinfo{year}{2020}\natexlab{}.
\newblock \showarticletitle{Incorporating user micro-behaviors and item
  knowledge into multi-task learning for session-based recommendation}. In
  \bibinfo{booktitle}{\emph{Proceedings of the 43rd international ACM SIGIR
  conference on research and development in Information Retrieval}}.
  \bibinfo{pages}{1091--1100}.
\newblock


\bibitem[Paperno(2016)]%
        {lambada-dataset}
\bibfield{author}{\bibinfo{person}{Denis et~al Paperno}.}
  \bibinfo{year}{2016}\natexlab{}.
\newblock \showarticletitle{The {LAMBADA} dataset: Word prediction requiring a
  broad discourse context}. In \bibinfo{booktitle}{\emph{Proceedings of the
  54th Annual Meeting of the Association for Computational Linguistics (Volume
  1: Long Papers)}}.
\newblock


\bibitem[Puy et~al\mbox{.}(2024)]%
        {Puy_2024_CVPR}
\bibfield{author}{\bibinfo{person}{Gilles Puy}, \bibinfo{person}{Spyros
  Gidaris}, \bibinfo{person}{Alexandre Boulch}, \bibinfo{person}{Oriane
  Sim\'eoni}, \bibinfo{person}{Corentin Sautier}, \bibinfo{person}{Patrick
  P\'erez}, \bibinfo{person}{Andrei Bursuc}, {and} \bibinfo{person}{Renaud
  Marlet}.} \bibinfo{year}{2024}\natexlab{}.
\newblock \showarticletitle{Three Pillars Improving Vision Foundation Model
  Distillation for Lidar}. In \bibinfo{booktitle}{\emph{Proceedings of the
  IEEE/CVF Conference on Computer Vision and Pattern Recognition (CVPR)}}.
  \bibinfo{pages}{21519--21529}.
\newblock


\bibitem[Radford et~al\mbox{.}(2019)]%
        {gpt2}
\bibfield{author}{\bibinfo{person}{Alec Radford}, \bibinfo{person}{Jeffrey Wu},
  \bibinfo{person}{Rewon Child}, \bibinfo{person}{David Luan},
  \bibinfo{person}{Dario Amodei}, \bibinfo{person}{Ilya Sutskever},
  {et~al\mbox{.}}} \bibinfo{year}{2019}\natexlab{}.
\newblock \showarticletitle{Language models are unsupervised multitask
  learners}.
\newblock \bibinfo{journal}{\emph{OpenAI blog}} \bibinfo{volume}{1},
  \bibinfo{number}{8} (\bibinfo{year}{2019}), \bibinfo{pages}{9}.
\newblock


\bibitem[Raffel et~al\mbox{.}(2020)]%
        {C4}
\bibfield{author}{\bibinfo{person}{Colin Raffel}, \bibinfo{person}{Noam
  Shazeer}, \bibinfo{person}{Adam Roberts}, \bibinfo{person}{Katherine Lee},
  \bibinfo{person}{Sharan Narang}, \bibinfo{person}{Michael Matena},
  \bibinfo{person}{Yanqi Zhou}, \bibinfo{person}{Wei Li}, {and}
  \bibinfo{person}{Peter~J. Liu}.} \bibinfo{year}{2020}\natexlab{}.
\newblock \showarticletitle{Exploring the Limits of Transfer Learning with a
  Unified Text-to-Text Transformer}.
\newblock \bibinfo{journal}{\emph{Journal of Machine Learning Research}}
  \bibinfo{volume}{21}, \bibinfo{number}{140} (\bibinfo{year}{2020}),
  \bibinfo{pages}{1--67}.
\newblock
\urldef\tempurl%
\url{http://jmlr.org/papers/v21/20-074.html}
\showURL{%
\tempurl}


\bibitem[Romero et~al\mbox{.}(2014)]%
        {romero2014fitnets}
\bibfield{author}{\bibinfo{person}{Adriana Romero}, \bibinfo{person}{Nicolas
  Ballas}, \bibinfo{person}{Samira~Ebrahimi Kahou}, \bibinfo{person}{Antoine
  Chassang}, \bibinfo{person}{Carlo Gatta}, {and} \bibinfo{person}{Yoshua
  Bengio}.} \bibinfo{year}{2014}\natexlab{}.
\newblock \showarticletitle{Fitnets: Hints for thin deep nets}.
\newblock \bibinfo{journal}{\emph{arXiv preprint arXiv:1412.6550}}
  (\bibinfo{year}{2014}).
\newblock


\bibitem[Shi et~al\mbox{.}(2023)]%
        {shi2023distributed}
\bibfield{author}{\bibinfo{person}{Hao-Jun~Michael Shi},
  \bibinfo{person}{Tsung-Hsien Lee}, \bibinfo{person}{Shintaro Iwasaki},
  \bibinfo{person}{Jose Gallego-Posada}, \bibinfo{person}{Zhijing Li},
  \bibinfo{person}{Kaushik Rangadurai}, \bibinfo{person}{Dheevatsa Mudigere},
  {and} \bibinfo{person}{Michael Rabbat}.} \bibinfo{year}{2023}\natexlab{}.
\newblock \showarticletitle{A distributed data-parallel pytorch implementation
  of the distributed shampoo optimizer for training neural networks at-scale}.
\newblock \bibinfo{journal}{\emph{arXiv preprint arXiv:2309.06497}}
  (\bibinfo{year}{2023}).
\newblock


\bibitem[Srinivasan et~al\mbox{.}(2022)]%
        {quill}
\bibfield{author}{\bibinfo{person}{Krishna Srinivasan},
  \bibinfo{person}{Karthik Raman}, \bibinfo{person}{Anupam Samanta},
  \bibinfo{person}{Lingrui Liao}, \bibinfo{person}{Luca Bertelli}, {and}
  \bibinfo{person}{Mike Bendersky}.} \bibinfo{year}{2022}\natexlab{}.
\newblock \bibinfo{title}{QUILL: Query Intent with Large Language Models using
  Retrieval Augmentation and Multi-stage Distillation}.
\newblock
\showeprint[arxiv]{2210.15718}~[cs.CL]
\urldef\tempurl%
\url{https://arxiv.org/abs/2210.15718}
\showURL{%
\tempurl}


\bibitem[Teerapittayanon et~al\mbox{.}(2016)]%
        {teerapittayanon2016branchynet}
\bibfield{author}{\bibinfo{person}{Surat Teerapittayanon},
  \bibinfo{person}{Bradley McDanel}, {and} \bibinfo{person}{Hsiang-Tsung
  Kung}.} \bibinfo{year}{2016}\natexlab{}.
\newblock \showarticletitle{Branchynet: Fast inference via early exiting from
  deep neural networks}. In \bibinfo{booktitle}{\emph{2016 23rd international
  conference on pattern recognition (ICPR)}}. IEEE,
  \bibinfo{pages}{2464--2469}.
\newblock


\bibitem[Touvron et~al\mbox{.}(2021)]%
        {touvron2021training}
\bibfield{author}{\bibinfo{person}{Hugo Touvron}, \bibinfo{person}{Matthieu
  Cord}, \bibinfo{person}{Matthijs Douze}, \bibinfo{person}{Francisco Massa},
  \bibinfo{person}{Alexandre Sablayrolles}, {and} \bibinfo{person}{Herv{\'e}
  J{\'e}gou}.} \bibinfo{year}{2021}\natexlab{}.
\newblock \showarticletitle{Training data-efficient image transformers \&
  distillation through attention}. In \bibinfo{booktitle}{\emph{International
  conference on machine learning}}. PMLR, \bibinfo{pages}{10347--10357}.
\newblock


\bibitem[Ullrich et~al\mbox{.}(2017)]%
        {ullrich2017soft}
\bibfield{author}{\bibinfo{person}{Karen Ullrich}, \bibinfo{person}{Edward
  Meeds}, {and} \bibinfo{person}{Max Welling}.}
  \bibinfo{year}{2017}\natexlab{}.
\newblock \showarticletitle{Soft weight-sharing for neural network
  compression}.
\newblock \bibinfo{journal}{\emph{arXiv preprint arXiv:1702.04008}}
  (\bibinfo{year}{2017}).
\newblock


\bibitem[Valipour et~al\mbox{.}(2023)]%
        {valipour2023sortednet}
\bibfield{author}{\bibinfo{person}{Mojtaba Valipour}, \bibinfo{person}{Mehdi
  Rezagholizadeh}, \bibinfo{person}{Hossein Rajabzadeh},
  \bibinfo{person}{Marzieh Tahaei}, \bibinfo{person}{Boxing Chen}, {and}
  \bibinfo{person}{Ali Ghodsi}.} \bibinfo{year}{2023}\natexlab{}.
\newblock \showarticletitle{Sortednet, a place for every network and every
  network in its place: Towards a generalized solution for training many-in-one
  neural networks}.
\newblock \bibinfo{journal}{\emph{arXiv preprint arXiv:2309.00255}}
  (\bibinfo{year}{2023}).
\newblock


\bibitem[Vaswani et~al\mbox{.}(2017)]%
        {transformer}
\bibfield{author}{\bibinfo{person}{Ashish Vaswani}, \bibinfo{person}{Noam
  Shazeer}, \bibinfo{person}{Niki Parmar}, \bibinfo{person}{Jakob Uszkoreit},
  \bibinfo{person}{Llion Jones}, \bibinfo{person}{Aidan~N Gomez},
  \bibinfo{person}{\L~ukasz Kaiser}, {and} \bibinfo{person}{Illia Polosukhin}.}
  \bibinfo{year}{2017}\natexlab{}.
\newblock \showarticletitle{Attention is All you Need}. In
  \bibinfo{booktitle}{\emph{Advances in Neural Information Processing
  Systems}}, \bibfield{editor}{\bibinfo{person}{I.~Guyon},
  \bibinfo{person}{U.~Von Luxburg}, \bibinfo{person}{S.~Bengio},
  \bibinfo{person}{H.~Wallach}, \bibinfo{person}{R.~Fergus},
  \bibinfo{person}{S.~Vishwanathan}, {and} \bibinfo{person}{R.~Garnett}}
  (Eds.), Vol.~\bibinfo{volume}{30}. \bibinfo{publisher}{Curran Associates,
  Inc.}
\newblock
\urldef\tempurl%
\url{https://proceedings.neurips.cc/paper_files/paper/2017/file/3f5ee243547dee91fbd053c1c4a845aa-Paper.pdf}
\showURL{%
\tempurl}


\bibitem[Vyas et~al\mbox{.}(2024)]%
        {vyas2024soap}
\bibfield{author}{\bibinfo{person}{Nikhil Vyas}, \bibinfo{person}{Depen
  Morwani}, \bibinfo{person}{Rosie Zhao}, \bibinfo{person}{Itai Shapira},
  \bibinfo{person}{David Brandfonbrener}, \bibinfo{person}{Lucas Janson}, {and}
  \bibinfo{person}{Sham Kakade}.} \bibinfo{year}{2024}\natexlab{}.
\newblock \showarticletitle{{SOAP}: Improving and stabilizing shampoo using
  {A}dam}.
\newblock \bibinfo{journal}{\emph{arXiv preprint arXiv:2409.11321}}
  (\bibinfo{year}{2024}).
\newblock


\bibitem[Wang(2024)]%
        {wang-etal-2024-train}
\bibfield{author}{\bibinfo{person}{Yueqi et~al. Wang}.}
  \bibinfo{year}{2024}\natexlab{}.
\newblock \showarticletitle{Train Once, Deploy Anywhere: Matryoshka
  Representation Learning for Multimodal Recommendation}. In
  \bibinfo{booktitle}{\emph{Findings of the Association for Computational
  Linguistics: EMNLP 2024}}. \bibinfo{publisher}{Association for Computational
  Linguistics}, \bibinfo{pages}{13461--13472}.
\newblock


\bibitem[Wu et~al\mbox{.}(2021)]%
        {wu2021one}
\bibfield{author}{\bibinfo{person}{Chuhan Wu}, \bibinfo{person}{Fangzhao Wu},
  {and} \bibinfo{person}{Yongfeng Huang}.} \bibinfo{year}{2021}\natexlab{}.
\newblock \showarticletitle{One teacher is enough? pre-trained language model
  distillation from multiple teachers}.
\newblock \bibinfo{journal}{\emph{arXiv preprint arXiv:2106.01023}}
  (\bibinfo{year}{2021}).
\newblock


\bibitem[Yang et~al\mbox{.}(2023)]%
        {yang2023does}
\bibfield{author}{\bibinfo{person}{Beining Yang}, \bibinfo{person}{Kai Wang},
  \bibinfo{person}{Qingyun Sun}, \bibinfo{person}{Cheng Ji},
  \bibinfo{person}{Xingcheng Fu}, \bibinfo{person}{Hao Tang},
  \bibinfo{person}{Yang You}, {and} \bibinfo{person}{Jianxin Li}.}
  \bibinfo{year}{2023}\natexlab{}.
\newblock \showarticletitle{Does graph distillation see like vision dataset
  counterpart?}
\newblock \bibinfo{journal}{\emph{Advances in Neural Information Processing
  Systems}}  \bibinfo{volume}{36} (\bibinfo{year}{2023}),
  \bibinfo{pages}{53201--53226}.
\newblock


\bibitem[Yu et~al\mbox{.}(2020)]%
        {yu2020dual}
\bibfield{author}{\bibinfo{person}{Jiahui Yu}, \bibinfo{person}{Wei Han},
  \bibinfo{person}{Anmol Gulati}, \bibinfo{person}{Chung-Cheng Chiu},
  \bibinfo{person}{Bo Li}, \bibinfo{person}{Tara~N Sainath},
  \bibinfo{person}{Yonghui Wu}, {and} \bibinfo{person}{Ruoming Pang}.}
  \bibinfo{year}{2020}\natexlab{}.
\newblock \showarticletitle{Dual-mode ASR: Unify and improve streaming ASR with
  full-context modeling}.
\newblock \bibinfo{journal}{\emph{arXiv preprint arXiv:2010.06030}}
  (\bibinfo{year}{2020}).
\newblock


\bibitem[Yuan et~al\mbox{.}(2020)]%
        {yuan2020parameter}
\bibfield{author}{\bibinfo{person}{Fajie Yuan}, \bibinfo{person}{Xiangnan He},
  \bibinfo{person}{Alexandros Karatzoglou}, {and} \bibinfo{person}{Liguang
  Zhang}.} \bibinfo{year}{2020}\natexlab{}.
\newblock \showarticletitle{Parameter-efficient transfer from sequential
  behaviors for user modeling and recommendation}. In
  \bibinfo{booktitle}{\emph{Proceedings of the 43rd International ACM SIGIR
  conference on research and development in Information Retrieval}}.
  \bibinfo{pages}{1469--1478}.
\newblock


\bibitem[Zellers et~al\mbox{.}(2019)]%
        {hellaswag}
\bibfield{author}{\bibinfo{person}{Rowan Zellers}, \bibinfo{person}{Ari
  Holtzman}, \bibinfo{person}{Yonatan Bisk}, \bibinfo{person}{Ali Farhadi},
  {and} \bibinfo{person}{Yejin Choi}.} \bibinfo{year}{2019}\natexlab{}.
\newblock \showarticletitle{HellaSwag: Can a Machine Really Finish Your
  Sentence?}. In \bibinfo{booktitle}{\emph{Proceedings of the 57th Annual
  Meeting of the Association for Computational Linguistics}}.
\newblock


\bibitem[Zhang et~al\mbox{.}(2019)]%
        {zhang2019your}
\bibfield{author}{\bibinfo{person}{Linfeng Zhang}, \bibinfo{person}{Jiebo
  Song}, \bibinfo{person}{Anni Gao}, \bibinfo{person}{Jingwei Chen},
  \bibinfo{person}{Chenglong Bao}, {and} \bibinfo{person}{Kaisheng Ma}.}
  \bibinfo{year}{2019}\natexlab{}.
\newblock \showarticletitle{Be your own teacher: Improve the performance of
  convolutional neural networks via self distillation}. In
  \bibinfo{booktitle}{\emph{Proceedings of the IEEE/CVF international
  conference on computer vision}}. \bibinfo{pages}{3713--3722}.
\newblock


\bibitem[Zhang et~al\mbox{.}(2018)]%
        {zhang2018deep}
\bibfield{author}{\bibinfo{person}{Ying Zhang}, \bibinfo{person}{Tao Xiang},
  \bibinfo{person}{Timothy~M Hospedales}, {and} \bibinfo{person}{Huchuan Lu}.}
  \bibinfo{year}{2018}\natexlab{}.
\newblock \showarticletitle{Deep mutual learning}. In
  \bibinfo{booktitle}{\emph{Proceedings of the IEEE conference on computer
  vision and pattern recognition}}. \bibinfo{pages}{4320--4328}.
\newblock


\bibitem[Zhong(2024)]%
        {sat-agieval}
\bibfield{author}{\bibinfo{person}{Wanjun et~al Zhong}.}
  \bibinfo{year}{2024}\natexlab{}.
\newblock \showarticletitle{{AGIE}val: A Human-Centric Benchmark for Evaluating
  Foundation Models}. In \bibinfo{booktitle}{\emph{Findings of the Association
  for Computational Linguistics: NAACL 2024}}. \bibinfo{publisher}{Association
  for Computational Linguistics}, \bibinfo{pages}{2299--2314}.
\newblock


\end{thebibliography}

\end{document}